\documentclass{article} % For LaTeX2e
\usepackage[preprint]{colm2026_conference}

\usepackage{microtype}
\usepackage{hyperref}
\usepackage{url}
\usepackage{booktabs}
\usepackage{graphicx}
\usepackage{array}

\usepackage{lineno}

\definecolor{darkblue}{rgb}{0, 0, 0.5}
\hypersetup{colorlinks=true, citecolor=darkblue, linkcolor=darkblue, urlcolor=darkblue}

\newcommand{\blue}[1]{\textcolor{black}{#1}}

\title{Phonological Perception of Sign Language Models}

\author{Kayo Yin\thanks{Correspondence: \texttt{kayoyin@berkeley.edu}} \\
University of California, Berkeley \\
\And
Jessica Carter \\
Johns Hopkins University \\
\AND
Alex X. Lu \\
Microsoft Research \\
\And
Annemarie Kocab \\
Johns Hopkins University \\
}

\newcolumntype{C}[1]{>{\centering\arraybackslash}m{#1}}

\begin{document}

\ifcolmsubmission
\linenumbers
\fi

\maketitle

\begin{abstract}
Sign languages are compositional systems where meaning arises by combining sublexical phonological parameters, such as handshape, location, and movement. While deep learning models for Sign Language Recognition (SLR) have achieved increased performance on translation benchmarks, it remains unclear whether these models distinguish abstract phonological features or merely rely on low-level statistical correlations. This work evaluates the phonological perception of SLR models trained on American Sign Language (ASL) by probing phonological sensitivity using minimal pairs and evaluating representational alignment with human behavioral data. Our results reveal that SLR models exhibit emergent phonological sensitivity, but with clear architectural trade-offs: pose-based models are sensitive to handshape contrasts, while pixel-based models better capture location changes. Furthermore, pose-based models learn latent representations that correlate with human perceptual similarity judgments ($r \approx0.49$). These findings suggest that while SLR models exhibit emergent phonology, current training paradigms are insufficient to scale them beyond their architectural inductive biases.\blue{\footnote{Code and data: \url{https://github.com/kayoyin/sign-phonology}.}}

\end{abstract}

\section{Introduction}

In signed languages, signs are constructed from sublexical phonological parameters: handshape, location, movement, orientation, and non-manual markers \citep{stokoe}. While modern Sign Language Recognition (SLR) models achieve high translation fidelity \citep{camgoz2020sign, yin-read-2020-better, guan2025mska}, it remains unclear if they acquire robust phonological representations. 
% \blue{To calibrate where SLR currently stands, isolated SLR models reach 63\% top-1 accuracy and 91\% recall-at-10 on ASL Citizen \citep{desai2023asl}, while recent continuous translation models reach BLEU-4 around 24 on RWTH-PHOENIX-2014T \red{[Lin et al., 2025 --- add citation to colm2026\_conference.bib]}.}
Current evaluations focus on translation accuracy on narrow datasets \citep{desai2024systemic}. However, high accuracy does not guarantee linguistic competence; it could mask a limited ability to generalize beyond training distributions. High scores may instead stem from overfitting to supplementary cues such as mouthing, which---unlike core phonological parameters---is optional and varies across signs and signers. The model may not generalize to the significant variability in signing across signers, or overfit to the recording conditions in the training data. This reliance on spurious features poses significant limitations for using SLR models as general-purpose tools for linguistic analysis, especially for low-resource sign languages. Given that distinguishing phonological parameters is critical for human lexical access \citep{lieberman2020lexical,williams2017operationalization}, if models are not sensitive to the true underlying phonology, they lack the robust representational quality required for cross-linguistic transfer or language understanding.

\begin{figure}[tbp]
    \centering
    \includegraphics[width=0.8\linewidth]{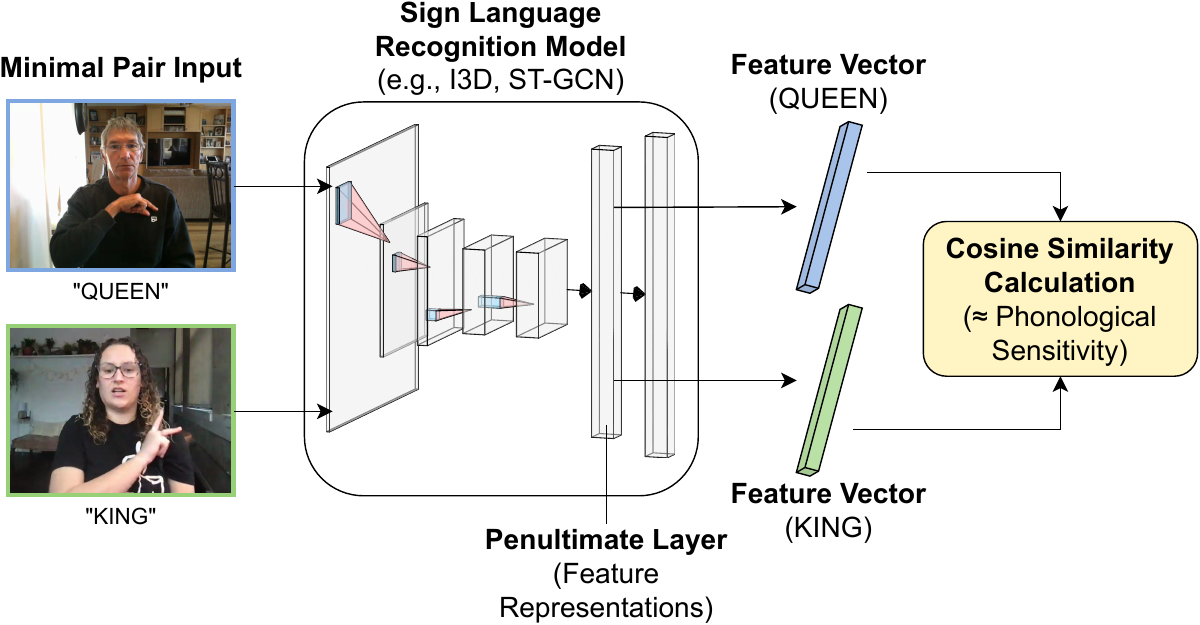}
\caption{\textbf{Auditing phonological sensitivity with minimal pairs}. Feature representations of each sign in a minimal pair (e.g. \textit{QUEEN} vs. \textit{KING}) are extracted from the model's penultimate layer. We quantify the model's sensitivity to the phonological contrast by calculating the cosine similarity between these latent representations.}
\label{fig:fig1}
\end{figure}

In this work, we evaluate whether SLR models exhibit phonological sensitivity by probing them with \textbf{minimal pairs }of signs (Figure \ref{fig:fig1}). These signs differ by one phonological element (e.g., \textit{QUEEN} vs. \textit{KING}, which share the same articulation location and sign movement and differ only in their handshape). This framework allows us to explicitly measure whether the model's latent space distinguishes signs along their phonological dimensions. Furthermore, by studying latent representations instead of model outputs, our framework directly compares various architectures and data sources without requiring explicit training on phonological labels.

We apply this framework in two experiments. First, we assess discriminability in naturalistic American Sign Language (ASL) data. We compare distinct architectures (pixel-based vs. pose-based) and training domains (sign language vs. general action recognition) to determine which features facilitate phonological sensitivity. 

Second, we investigate the alignment between model representations and human perception. We use controlled synthetic data to study how the latent space of models organize handshapes compares to theoretical models of phonology based on human perception data.

Our results reveal trade-offs in model sensitivity governed by architectural constraints. While pixel-based models better capture broader spatial contrasts like sign location, pose-based models demonstrate superior sensitivity to fine-grained handshape distinctions. We also find that the latent space of pose-based models significantly aligns with human perceptual judgments, partially reproducing the confusion patterns and hierarchical groupings found in human signers. These findings suggest that while emergent phonology exists in SLR models, models designed with different visual processing show different phonological sensitivities, highlighting the critical importance of cognitively grounded benchmarks to guide the development of robust computational sign systems.

\section{Background \& Related Work}

\paragraph{Phonological structure of sign languages.}
Sign languages contain sublexical structure where meaningful signs are constructed from combinatorial parameters. We adopt the standard five-parameter model established by \citet{battison1978lexical} and \citet{liddell1989american}, which expands \citet{stokoe}'s original classification to include: \textbf{handshape}, \textbf{location}, \textbf{movement}, \textbf{palm orientation}, and \textbf{non-manual markers}. This compositional structure allows us to define \textit{minimal pairs} -- signs that differ by exactly one parameter -- to measure discriminability. While originally described for ASL, this parameter model also generalizes across different sign languages \citep{sandler2012phonological}. Psycholinguistic research demonstrates that signers activate these sublexical representations incrementally during lexical access \citep{meade2018phonological}, while behavioral studies show that phonological similarity influences sign production and recognition speeds \citep{carreiras2008lexical, williams2017operationalization}.

\paragraph{Sign language recognition.}
SLR research has largely prioritized end-to-end translation accuracy, often treating models as ``black boxes" whose internal representations remain unexamined \citep{camgoz2020sign, yin-read-2020-better}. Existing attempts to analyze phonology typically rely on explicit supervision, training classifiers to predict specific features \citep{tavella2022phonology, bilge2022towards}. However, this approach is limited by data scarcity for rare features and statistical confounds (e.g., specific handshapes correlate with specific locations, so a classifier can score well by predicting a feature whenever a correlated parameter occurs, rather than by detecting the parameter of interest). To avoid these validity issues, we assess \textit{implicit} phonological emergence using a minimal pair strategy. This method controls for confounding variables and requires no phonological labels, allowing us to probe the model's latent structure directly.

% Research in sign language processing has largely prioritized end-to-end translation tasks, using architectures like 3D-CNNs or Transformers to map video to text \citep{camgoz2020sign, yin-read-2020-better, guan2025mska}. While these models achieve high benchmark performance, the properties of their latent representations are rarely analyzed. Furthermore, the predominance of translation tasks reflects a systemic bias that views sign language primarily as an accessibility tool rather than a linguistic system in its own right \citep{desai2024systemic}.

% Existing work involving sign phonology typically relies on explicit supervision, either predicting phonological properties \citep{tavella2022phonology, kezar2023exploring} or using them as intermediate training signals \citep{bilge2022towards, kezar2023sem}. To probe phonological representations, one might train a classifier to predict phonological features. However, this approach faces two major validity issues: (1) correlations between parameters (e.g., certain handshapes co-occurring only with specific locations) can confound results, and (2) rare phonological features often lack sufficient training data. In contrast, we investigate whether phonological distinctions emerge implicitly in models trained without phonological labels. We employ a minimal pair evaluation strategy -- this approach is robust to data scarcity and strictly controls for confounding variables, isolating specific phonological changes.

\paragraph{Phonological representations in spoken language models.}

Analogous work in speech processing demonstrates that in neural networks, phonological structure is emergent without explicit supervision. Research shows that phonological information is encoded in the lower to middle layers of speech recognition models \citep{belinkov2017analyzing, pasad2021layer} and can be retrieved via linear geometry \citep{gauthier2025emergent}. We extend this line of inquiry to the visual-spatial domain, investigating whether sign language models exhibit similar emergent sensitivity to phonological constraints.

% While the analysis of learned phonology in sign language models is relatively nascent, there has been extensive analogous work in speech models. \citet{belinkov2017analyzing} demonstrated that LSTM-based speech recognition systems encode phonological information in lower layers. Recent studies on self-supervised models \citep{pasad2021layer, chung2021similarity} located phonology in the middle layers, noting that masked prediction objectives yield more discrete clustering than contrastive objectives. Additionally, \citet{gauthier2025emergent} found that these models can encode phonological rules through continuous linear representations. We extend this inquiry to the visual-spatial domain, analyzing whether sign language models exhibit similar emergent phonological sensitivity.

\section{Methodology}

To assess the phonological sensitivity of SLR models, we use a \textbf{minimal pair} framework. Rather than relying on final classification outputs, we analyze the models' latent representations of sign pairs that differ by a single phonological parameter. This approach allows us to quantify sensitivity to specific phonological changes in models trained without explicit phonological supervision.

\subsection{Minimal pairs and datasets}
\begin{figure}[]
    \centering
    \includegraphics[width=0.7\linewidth]{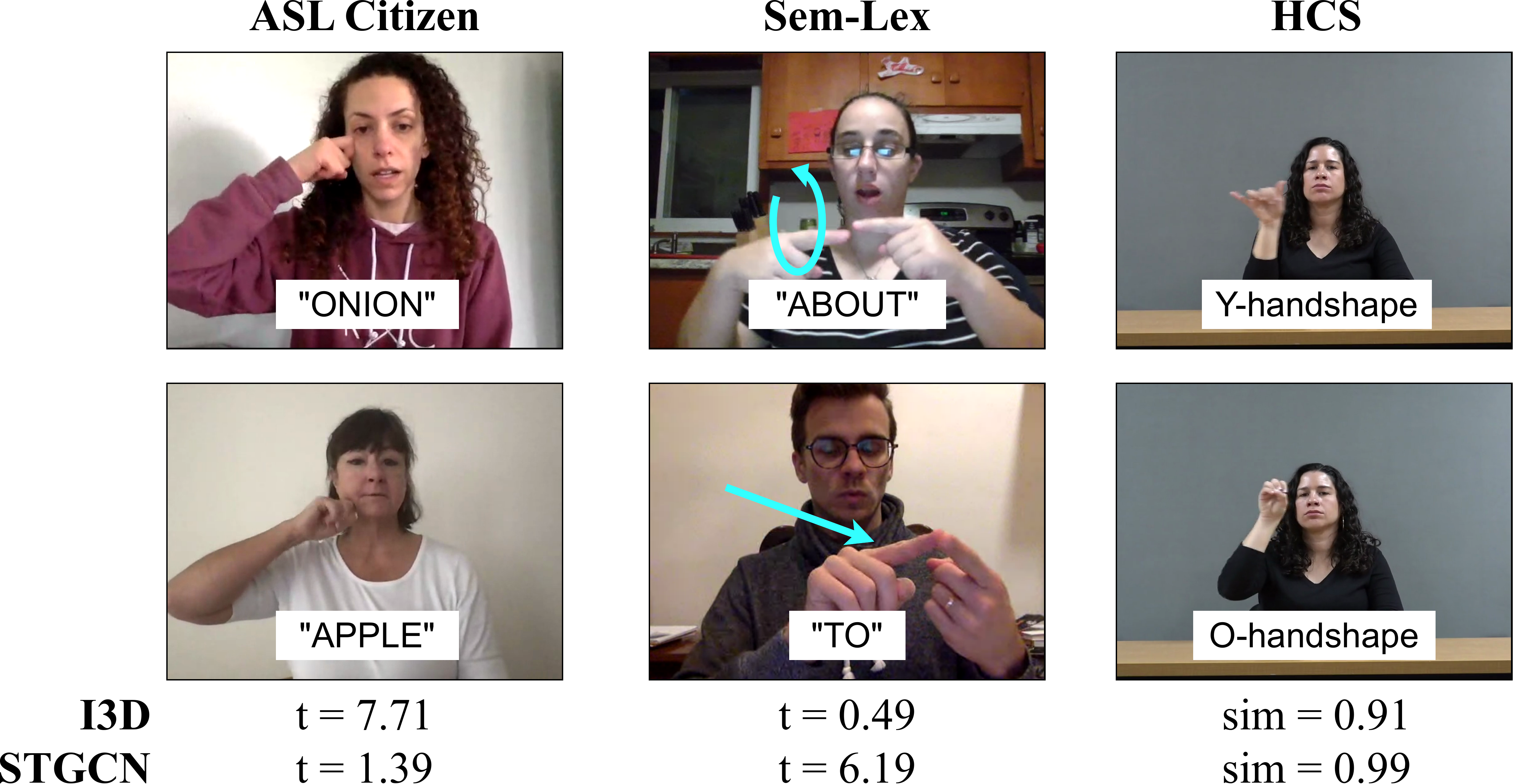}
\caption{\textbf{Examples of minimal pair data and model sensitivity metrics.} (Left, Center) Naturalistic minimal pairs from ASL Citizen and Sem-Lex contrasting in Handshape and Location, respectively. Below: $t$-test results indicate the statistical significance of SLR models' ability to distinguish these pairs. (Right) A controlled minimal pair from HCS contrasting Handshape. Below: Cosine similarity scores quantify the distance between the two nonce signs in the models' latent space (lower similarity = higher sensitivity).}
\label{fig:data}
\end{figure}

We curate minimal pairs from three datasets: ASL Citizen \citep{desai2023asl}, Sem-Lex \citep{kezar2023sem}, and Handshapes in Context Stimuli \citep{hcs}. We adopt the parameter inventory of \citet{battison1978lexical} and \citet{liddell1989american} for minimal-pair decisions: each minimal pair contains exactly one change in handshape, location, movement, orientation, or non-manual markers, verified by a Deaf author. Our experiments focus on handshape, location, and movement contrasts, as orientation and non-manual contrasts are too sparse in the available naturalistic corpora to evaluate reliably. We provide samples from each dataset in Figure \ref{fig:data}.

% locations = ['Elbow', 'Mouth', 'Neutral']
% movements = ['Circle', 'Lateral', 'Twist']
% orientations = ['Down', 'Forward', 'Self']

\paragraph{ASL Citizen.} ASL Citizen is a large-scale, crowd-sourced dataset featuring 52 deaf or hard-of-hearing (DHH) signers. A Deaf author curated 259 minimal pairs spanning 360 unique signs to cover diverse phonological changes: \blue{137} pairs differ in handshape, 31 in location, \blue{73} in movement, and \blue{18} in orientation or non-manual markers. To establish a baseline for human perceptual similarity, we enlisted a hearing \blue{novice} signer to annotate the pairs as ``very similar'' (51 pairs), ``somewhat similar'' (98 pairs), or ``not similar'' (103 pairs). We use the standard test split, which includes 4,427 videos from 11 signers for the selected signs.

\paragraph{Sem-Lex.} As all SLR models in this study were trained on ASL Citizen, to test for generalization, we also curated pairs from the Sem-Lex Benchmark. Unlike ASL Citizen, where users mimicked a seed video, Sem-Lex prompted signers with concepts, which resulted in natural variability in sign production. We mapped our ASL Citizen minimal pairs to Sem-Lex by manually verifying that each video sharing a gloss matched the intended sign, and excluding lexical variants whose production differed from its pair partner in more than one parameter. This ensures our framework evaluates robust phonological sensitivity rather than being confounded by mislabeled videos or lexical variation. This yielded 178 pairs across 263 signs (8,934 videos from 42 signers).

\paragraph{Handshapes in Context Stimuli (HCS).} Real-world vocabularies often lack systematic coverage of phonological combinations. To address this, we use HCS, a set of controlled experimental stimuli consisting of nonce signs designed to systematically sample variations in handshape and phonological context. The design of these stimuli was guided by prior phonological inventories \citep{lane1976preliminaries, stungis}, while the specific handshape inventory was expanded from 20 to 36 shapes based on frequency scores from a lexical database \citep{asl-lex}. HCS includes 972 videos covering 36 handshapes in 27 phonological contexts (3 locations × 3 movements × 3 orientations). This systematic design allows for precise minimal pair contrasts even for features rare in natural vocabulary. Unlike ASL Citizen and Sem-Lex, HCS features one Deaf signer with one video per nonce sign, recorded in a controlled laboratory setting. 

\subsection{Models}

We evaluate variants of two distinct video classification architectures, I3D and STGCN, which represent the predominant modeling paradigms \footnote{ I3D and STGCN are the dominant open-weight representatives of pixel- and pose-based SLR at the time of publication. Although more advanced systems have been proposed in more recent work, these largely do not make their weights openly available, which precludes systematic comparison.} in the current SLR literature (i.e. vision versus pose-based) \citep{desai2023asl}.
By comparing these standard backbones, we investigate how \textit{training domain} (general human action vs. sign language recognition) and \textit{input modality} (raw pixels vs. pose estimations) influence phonological perception. The two input modalities are illustrated in Appendix \ref{app:input}.

\paragraph{I3D (Pixel-based).} The Inflated 3D ConvNet (I3D; \citet{carreira2017quo}) is a two-stream architecture operating directly on RGB video frames. We evaluate three variants: (1) \textbf{I3D-Rand}, initialized with random weights to serve as a baseline; (2) \textbf{I3D-Kine}, trained on the Kinetics dataset for general action recognition; and (3) \textbf{I3D-ASL}, trained on ASL Citizen for sign language recognition.

\paragraph{STGCN (Pose-based).} The Spatio-Temporal Graph Convolutional Network (STGCN; \citet{yan2018spatial}) operates on a skeletal graph representation of the body. We evaluate: (1) \textbf{STGCN-Rand}, a randomly initialized baseline; and (2) \textbf{STGCN-ASL}, trained on ASL Citizen. We excluded an STGCN model trained on Kinetics because existing pre-trained models use action recognition pose graphs that lack fine-grained hand landmarks.

\paragraph{Strict exclusion of test signs.} To ensure our analysis measures generalized phonological sensitivity rather than lexical memorization, we re-trained both ASL-trained models (I3D-ASL and STGCN-ASL) on a dataset split that explicitly excludes all signs appearing in our minimal pair test sets.

\begin{table*}[t]
    \centering
    \caption{\textbf{Model sensitivity to phonological changes on ASL Citizen.} Results are grouped by linguistic feature (Handshape, Location, Movement) and human perception of similarity. $t$-stat represents the mean $t$-statistic calculated across all minimal pairs in that category. \% wins (vs. Baseline) denotes the percentage of pairs where the trained model had a significant $t$-statistic ($p<0.05$) that was higher than its randomly initialized counterpart. Head-to-Head columns show the percentage of pairs where one trained model (I3D-ASL or STGCN-ASL) significantly outperformed the other ($p<0.05$). Note that Head-to-Head percentages do not sum to 100\% as pairs with no significant $t$-stat are excluded.
    STGCN is generally more sensitive to phonological changes than I3D, except for changes in location.}
    \vspace{10pt}
    \resizebox{\linewidth}{!}{
\begin{tabular}{lc|cc|cc|c|cc|cc}
\toprule
 & \multicolumn{1}{c|}{\textbf{Baseline}} & \multicolumn{4}{c|}{\textbf{Pixel-based (I3D)}} & \multicolumn{1}{c|}{\textbf{Baseline}} & \multicolumn{2}{c|}{\textbf{Pose-based (STGCN)}} & \multicolumn{2}{c}{\textbf{Head-to-Head}} \\
\cmidrule(lr){2-2} \cmidrule(lr){3-6} \cmidrule(lr){7-7} \cmidrule(lr){8-9} \cmidrule(lr){10-11}
 & I3D-Rand & \multicolumn{2}{c|}{I3D-Kinetics} & \multicolumn{2}{c|}{\textbf{I3D-ASL}} & STGCN-Rand & \multicolumn{2}{c|}{\textbf{STGCN-ASL}} & \textbf{I3D-ASL} & \textbf{STGCN-ASL} \\
\textbf{Category} & $t$-stat & $t$-stat & \% wins & $t$-stat & \% wins& $t$-stat & $t$-stat & \% wins & \% wins & \% wins \\
\midrule
\textit{Total ($N = 259$)} & 0.13 & 0.12 & 1.93 & 5.00 & 69.50 & 0.23 & \textbf{6.08} & \textbf{81.08} & 33.20 & \textbf{56.37} \\
\blue{\textit{Control ($N = 259$)}} & \blue{0.06} & \blue{0.76} & \blue{---} & \blue{\textbf{10.98}} & \blue{---} & \blue{0.44} & \blue{\textbf{12.66}} & \blue{---} & \blue{---} & \blue{---} \\
\midrule
\multicolumn{11}{l}{\textit{By Phonological Parameter}} \\
Handshape ($N = 137$) & 0.14 & 0.12 & 0.73 & 4.80 & 69.34 & 0.18 & \textbf{6.30} & \textbf{86.13} & 29.20 & \textbf{64.96} \\
Location ($N = 31$) & 0.02 & 1.13 & 6.45 & \textbf{7.73} & \textbf{80.65} & 0.59 & 5.67 & 67.74 & \textbf{64.52} & 16.13 \\
Movement ($N = 73$) & 0.19 & 0.20 & 1.37 & 4.72 & 73.97 & 0.30 & \textbf{5.81} & \textbf{80.82} & 38.36 & \textbf{54.79} \\
\midrule
\multicolumn{11}{l}{\textit{By Human Perception}} \\
Not similar ($N = 103$) & 0.15 & 0.35 & 0.97 & 5.01 & 66.99 & 0.10 & \textbf{6.98} & \textbf{89.32} & 28.16 & \textbf{64.08} \\
Somewhat similar ($N = 98$) & 0.17 & 0.06 & 2.04 & 5.08 & 76.53 & 0.43 & \textbf{5.64} & \textbf{79.59} & 37.76 & \textbf{56.12} \\
Very similar ($N = 51$) & 0.07 & 0.17 & 3.92 & 4.26 & 56.86 & 0.03 & \textbf{4.62} & \textbf{64.71} & 31.37 & \textbf{43.14} \\
\bottomrule
    \end{tabular}}
    \label{tab:citizen}
\end{table*}

\begin{table}[ht]
    \centering
    \caption{\textbf{Out-of-domain phonological sensitivity results on Sem-Lex.}  Models trained on ASL Citizen were evaluated on minimal pairs from the Sem-Lex dataset to test domain generalization. Both models demonstrate a reduced sensitivity to phonological differences in this out-of-domain setting compared to ASL Citizen.}
    \vspace{10pt}
\resizebox{0.7\linewidth}{!}{
\begin{tabular}{l cc | cc | cc}
\toprule
%  & \multicolumn{2}{c|}{} & \multicolumn{2}{c|}{} & \multicolumn{2}{c}{\textbf{Head-to-Head}} \\
% \cmidrule(lr){2-3} \cmidrule(lr){4-5} \cmidrule(lr){6-7}
 & \multicolumn{2}{c|}{\textbf{I3D-ASL}} & \multicolumn{2}{c|}{\textbf{STGCN-ASL}} & \multicolumn{2}{c}{\textbf{Head-to-Head}} \\
\textbf{Category} & $t$-stat & \% wins & $t$-stat & \% wins & I3D & STGCN \\
\midrule
\textit{Total ($N = 71$)} & 2.47 & 45.07 & \textbf{2.76} & \textbf{49.30} & 25.35 & \textbf{33.80} \\
\blue{\textit{Control ($N = 71$)}} & \blue{\textbf{8.39}} & \blue{---} & \blue{\textbf{10.40}} & \blue{---} & \blue{---} & \blue{---} \\
\midrule
\multicolumn{7}{l}{\textit{By Phonological Parameter}} \\
Handshape ($N = 26$) & 3.29 & 57.69 & \textbf{3.93} & \textbf{61.54} & 30.77 & \textbf{38.46} \\
Location ($N = 8$) & 3.52 & \textbf{75.00} & \textbf{4.79} & 62.50 & 25.00 & \textbf{50.00} \\
Movement ($N = 25$) & \textbf{2.47} & 36.00 & 1.63 & \textbf{48.00} & 32.00 & 32.00 \\
\midrule
\multicolumn{7}{l}{\textit{By Human Perception}} \\
Not similar ($N = 23$) & \textbf{3.92} & \textbf{60.87} & 3.07 & 56.52 & 34.78 & \textbf{39.13} \\
Somewhat sim. ($N = 31$) & 2.05 & 41.94 & \textbf{2.39} & \textbf{45.16} & \textbf{29.03} & 25.81 \\
Very similar ($N = 17$) & 1.29 & 29.41 & \textbf{2.99} & \textbf{47.06} & 5.88 & \textbf{41.18} \\
\bottomrule
\end{tabular}}
 \label{tab:semlex}
\end{table}

\section{Are Models Sensitive to Phonological Changes?}

\subsection{Experiment}

We evaluate whether SLR models learn latent representations that preserve the phonological distinctions found in minimal pairs. By focusing on the latent feature space rather than final prediction accuracy, we establish a unified evaluation framework that allows for direct comparison across diverse model architectures and training objectives. Furthermore, we posit that robust phonological separability in the latent space is a strong indicator of a model's capacity for zero-shot transfer—a critical capability for analyzing underrepresented sign languages where training data are scarce.

To analyze these representations, we extract feature vectors from the model's penultimate layer (preceding the classification head) for videos in our naturalistic minimal pair datasets, where there are multiple samples per sign (ASL Citizen and Sem-Lex). We operationalize ``phonological sensitivity'' by comparing intra-sign similarity against inter-sign similarity. Our hypothesis is that if a model is sensitive to phonological change, the cosine similarity between two \textit{minimally different} signs should be \textbf{significantly lower} than two instances of the \textit{same} sign (produced by different signers).

Formally, for each minimal pair of signs $(s_1, s_2)$ we construct two distributions of similarity scores. \textbf{(1) Intra-sign similarity ($D_{same}$):} The cosine similarity between the representation of sign $s_1$ produced by signer $A$ and an instance of the same sign $s_1$ produced by a different signer $B$. \textbf{(2) Inter-sign similarity ($D_{diff}$):} The cosine similarity between the representation of sign $s_1$ produced by signer $A$ and the phonologically distinct sign $s_2$ produced by signer $B$. \textbf{(3) Sampling:} We repeat these computations for 10 random pairs of signers $(A, B)$ to populate the distributions $D_{same}$ and $D_{diff}$.

% \begin{enumerate}
%     \item \textbf{Intra-sign similarity ($D_{same}$):} The cosine similarity between the representation of sign $s_1$ produced by signer $A$ and an instance of the same sign $s_1$ produced by a different signer $B$.
%     \item \textbf{Inter-sign similarity ($D_{diff}$):} The cosine similarity between the representation of sign $s_1$ produced by signer $A$ and the phonologically distinct sign $s_2$ produced by signer $B$.
%     \item \textbf{Sampling:} We repeat these computations for 10 random pairs of signers $(A, B)$ to populate the distributions $D_{same}$ and $D_{diff}$.
% \end{enumerate}

To quantify phonological sensitivity, we perform a two-tailed independent samples t-test to determine if the distribution of similarities of $D_{same}$ is statistically significantly higher than that of $D_{diff}$\blue{:}

\begin{equation}
t(s_1, s_2) \;=\; \frac{\bar{D}_{same} - \bar{D}_{diff}}{\sqrt{s^2_{same}/N + s^2_{diff}/N}},
\label{eq:phon-sens}
\end{equation}
where $\bar{D}$ and $s^2$ denote the sample mean and variance of each similarity distribution, and $N$ is the number of sampled signer pairs.

As a control, we repeat this procedure on random non-minimal sign pairs sampled from the same gloss pool, providing an upper bound on the $t$-statistics that trained models should trivially achieve (\textit{Control} rows in Tables \ref{tab:citizen}--\ref{tab:semlex}).

\subsection{Results}

% \ky{TODO: add results showing that the models still fail to capture very obvious differences (not similar pairs) and both have strong failure modes: pick pairs that one model captures from not similar but not the other (so not a limitation of data) and show that some of the examples are completely missed by the models. Show certain frames, say we were surprised the number of times the model won against random is so low because they look very different}

We present the results of $t$-tests on ASL Citizen and Sem-Lex in Table \ref{tab:citizen} and Table \ref{tab:semlex}, respectively. For each phonological category, we report the \textbf{mean t-statistic} (magnitude of separation) and the \textbf{\% wins} (frequency with which the trained model significantly outperforms the randomly initialized baseline, $p<0.05$). Additionally, we provide head-to-head comparisons indicating how often one ASL-trained model proved significantly more sensitive than the other. In Appendix \ref{app:quant}, we also provide qualitative examples of model sensitivity results to minimal pairs in ASL-Citizen.

First, \textbf{training on sign language data is prerequisite for phonological sensitivity,} which notably deviates from human learning. Across all categories, models trained on general action recognition (I3D-Kinetics) failed to distinguish minimal pairs, achieving a negligible 1.93\% win rate over random baselines on ASL Citizen. In contrast, both ASL-trained models demonstrated significant emergent sensitivity (I3D-ASL: 69.50\%; STGCN-ASL: 81.08\%). The \textit{Control} row confirms that the ASL-trained models separate random non-minimal pairs far more strongly than minimal pairs (e.g., $t=12.66$ vs.\ $6.08$ for STGCN-ASL), while the untrained and Kinetics baselines remain at chance.

This failure of the general action model is surprising given that phonological sensitivity builds upon general visual capabilities: prior work demonstrates that non-signers can reliably distinguish phonological contrasts in ASL and there is a strong correlation between signer and non-signer perceptual confusion \citep{stungis}. Our results suggest that current video models are poor proxies for human perception. Unlike the human visual system, which can transfer general object and motion recognition to identify sign contrasts, these models require domain-specific training to acquire phonological sensitivity. We attribute this to the training signal itself: general action datasets like Kinetics lack the fine-grained phonological contrasts found in sign language, and so never pressure the model to encode the subtle distinctions that sign-specific training captures.

Second, \textbf{architectures exhibit distinct biases in what aspects of phonology they are sensitive to.} On ASL Citizen, the pose-based STGCN-ASL model demonstrates superior sensitivity to \textbf{handshape} over I3D-ASL (86.13\% vs. 69.34\%) and \textbf{movement} (80.82\% vs. 73.97\%). This suggests that explicit skeletal modeling better captures fine-grained configuration and trajectory than raw pixels. Conversely, the pixel-based I3D-ASL model significantly outperformed STGCN-ASL on \textbf{location} contrasts (80.65\% vs. 67.74\%), winning the head-to-head comparison in 64.52\% of location pairs. This indicates that pixel-based models may better retain the spatial context relative to the frame (e.g., forehead vs. chin) that graph-based representations -- which are often normalized for position -- can lose.

Third, \textbf{model sensitivity aligns with human perception.} For both architectures, performance degrades for minimal pairs rated by a human non-signer as perceptually more similar. A Kruskal-Wallis H-test confirms a statistically significant difference in STGCN-ASL sensitivity across the ``Not / Somewhat / Very Similar'' human ratings ($H(2)=12.92, p=0.002$), validating that the models struggle most with the distinctions humans may find subtle. In contrast, this difference was not statistically significant for the I3D-ASL model or the non-ASL trained variants.

% Fourth, \textbf{qualitative analysis reveals failure modes on seemingly ``obvious'' contrasts.} As the example shown in Figure \ref{fig:data}, models surprisingly fail to distinguish many minimal pairs rated as ``Not Similar.''. Crucially, these failures are often model-specific rather than the result of a data artifact.

Finally, \textbf{representations are brittle to domain shifts.} When evaluated on Sem-Lex, which introduces variations in prompts and filming conditions, sensitivity drops substantially (I3D-ASL: 45.07\%; STGCN-ASL: 49.3\%). This suggests that while current models learn phonological distinctions, their representations remain highly brittle to the specific production constraints of their training distribution.

\section{How Does Model Phonological Sensitivity Compare to Human Perception?}

\subsection{Experiment}

Beyond assessing whether models \textit{can} distinguish phonological features, we investigate whether their internal representation space captures meaningful phonological structure. We use the HCS data to compare model feature distances with subjective human perception and articulatory geometry. While naturalistic datasets like ASL Citizen offer ecological validity, they suffer from sparse phonological coverage. HCS allows us to control for non-phonological variance (e.g., lighting, signer identity) while systematically sampling the full spectrum of handshape contrasts.

First, we quantify \textbf{human perception} using confusion matrices collected by \citet{stungis}. This data quantifies how frequently human signers and non-signers confuse different ASL handshapes. We treat the confusion frequency as a proxy for perceptual similarity: handshapes that are confused more often are considered more perceptually similar.

Second, we calculate the \textbf{handshape distance (HD)} between handshape pairs. HD, a computational measure of articulatory geometry adapted from \citet{yin24acl}, is defined as the mean angular difference between their corresponding joints. For each pair of handshapes in HCS data, we computed the HD for the handshape pair in 27 phonological contexts and took the median.

Third, we measure \textbf{model sensitivity} to the phonological contrast in each pair of handshapes in HCS by calculating the cosine similarity between their corresponding video representations in the model's latent space. We then compute the Pearson correlation coefficient ($r$) between the vector of human confusion frequencies, handshape distances, and model feature similarities. We summarize the results of our correlation analysis in Table \ref{tab:cm}. 

\begin{figure}[]
    \centering
    \includegraphics[width=0.9\linewidth]{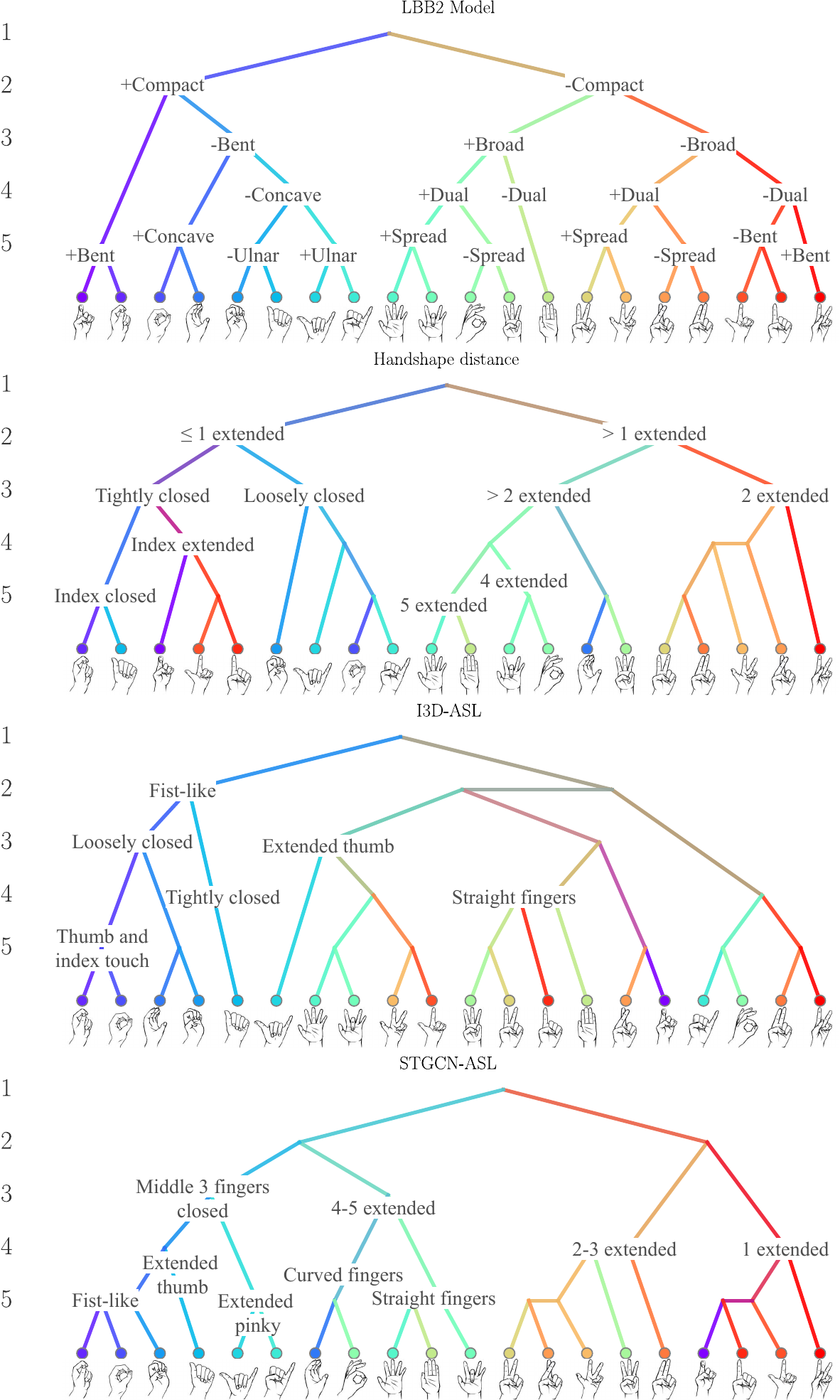}
\caption{\textbf{Structural organization of handshape representations.} From top to bottom: the LBB2 Model (theoretic model based on human perception), handshape distance (geometric reference), I3D-ASL, STGCN-ASL.}
\label{fig:tree}
\end{figure}

To visualize the structural organization of the learned feature spaces, we performed U-statistic hierarchical clustering \citep{d1978u} of handshapes according to HD and model feature similarity, following the methodology in \citet{stungis}. In Figure \ref{fig:tree}, we plotted dendrograms for four representations: (1) the \textbf{Lane-Boyes-Braem II (LBB2) Model} (a theoretical model based on human perceptual studies \citep{lane1976preliminaries, stungis}), (2) \textbf{Handshape Distance }(our geometric reference metric), (3) \textbf{I3D-ASL}, and (4)\textbf{ STGCN-ASL}. This allows us to inspect whether models recover high-level phonological features that humans rely on, or if they use different features for distinguishing handshapes.

% To do this, we compare model-derived feature similarities against human confusion matrices from \citet{stungis}. In this framework, we operationalize human confusion frequency as a proxy for perceptual similarity: handshapes that humans frequently confuse are considered perceptually close.

% While naturalistic datasets like ASL Citizen provide high real-world validity, they are limited by sparse phonological coverage. Natural vocabularies do not systematically sample the full combinatorial space of phonological features. To address this gap and control for non-phonological variance (e.g., lighting, signer identity), we use the HCS dataset, which provides a dense, systematic sampling of the full spectrum of handshape contrasts.

% For each pair of handshapes in the HCS dataset, we calculate the cosine similarity between their corresponding video representations in the model's latent space. We then compute the Pearson correlation coefficient ($r$) between the vector of model similarities and the vector of human confusion frequencies. A significant positive correlation indicates that the model's learned representations reflects human-like perceptual biases, grouping handshapes together that humans also find visually similar.

\begin{table}[htp]
    \centering
    \caption{\textbf{Correlation between perceptual similarity reported by humans \citep{stungis}, handshape distance (HD), and cosine similarity of model features of handshapes.} Significant correlations ($p < 0.05$) are bolded. Models trained on ASL data, especially STGCN, have higher alignment with human perception.}
    \vspace{10pt}
        \resizebox{0.8\linewidth}{!}{
    \begin{tabular}{lccc | ccc| cc}
    \toprule
     & \multicolumn{3}{c|}{\textbf{Reference}}  &  \multicolumn{3}{c|}{\textbf{I3D}} &  \multicolumn{2}{c}{\textbf{STGCN}} \\
\cmidrule(lr){2-4} \cmidrule(lr){5-7} \cmidrule(lr){8-9} 
    & Non-signer & Signer & HD & Rand & Kine & ASL & Rand & ASL \\
    \midrule
    Non-signer & -- & \textbf{0.89} & \textbf{0.48} & 0.04 & \textbf{0.19} & \textbf{0.32} & 0.10 & \textbf{0.48}\\
    Signer & \textbf{0.89} & -- & \textbf{0.53} & 0.03 & \textbf{0.19} & \textbf{0.31} & 0.14 & \textbf{0.49} \\
    HD & \textbf{0.48} & \textbf{0.53} & -- & 0.10 & \textbf{0.19} & \textbf{0.20} & 0.06 & \textbf{0.55} \\
     \bottomrule
    \end{tabular}}
    \label{tab:cm}
\end{table}

\subsection{Results}

\paragraph{Correlation analysis.}
First, we established reference bounds for our analysis. Consistent with prior literature, human signers and non-signers have high agreement ($r=0.89$). The theoretical HD metric showed moderate positive correlation with human perception, suggesting that human perception is grounded in articulatory geometry ($r=0.53$ for signers). As expected, models with randomly initialized weights (\textit{Rand}) yielded insignificant correlations, confirming that untrained networks do not inherently encode phonologically meaningful features.

Second, we find that \textbf{training on ASL data significantly improved alignment with human perception}. For the I3D architecture, the model trained on general human action recognition (\textit{Kine}) has a weak correlation with human signers ($r=0.19$). Training on ASL data increased this correlation to $r=0.31$ for the I3D model, suggesting that exposure to sign language data drives the feature space closer to human perceptual representations.

Third, we observe clear \textbf{architectural differences in handshape perception} between models. STGCN-ASL is the model with the highest alignment with human signers ($r=0.49$). The correlation between I3D-ASL and STGCN-ASL is itself notably low ($r=0.19$). This indicates that while both models successfully distinguish some handshapes, their architectures induce different structures in their latent spaces. STGCN-ASL also shows a strong correlation with handshape distance ($r=0.55$), substantially higher than I3D-ASL ($r=0.20$). I3D, lacking the skeletal prior of pose-based models, seems to learn a visual abstraction that correlates less with joint geometry and the human perception of handshape. We also note that the alignment between STGCN-ASL and humans is not higher than the alignment between HD and humans, which may suggest that training on sign language allows for similarities already present in input space to trickle through the model, but does not enhance perceptual similarity past what is expected from joint geometry.

% \begin{figure}[]
%     \includegraphics[width=\linewidth]{figures/cm.pdf}
% \caption{\textbf{Correlation between human perceptual judgments, representation similarity of SLR models, and handshape distance.} Points represent specific pairs of handshapes. To visualize structural alignment, points are color-coded by interpolating the colors assigned to the handshapes in the Figure \ref{fig:tree} dendrogram; saturation and opacity are scaled by the tree distance between the pair, such that perceptually distant handshapes appear less saturated. \ky{Remove scatter plots and make table of correlations instead}}
% \label{fig:cm}
% \end{figure}

% \ky{TODO: add illustrations of each handshape below leaves. Also, add the dendrograms obtained by hierarchical clustering of model similarities, handshape distance? Also replace handshape names with the modern ones, add footnote to explain difference with stungis names}
    % 'Closed_5': 'B',
    % '1': 'G',
    % 'Open_8': '8',
    % 'U': 'H',

% Finally, we compare the handshape phonological structure induced by different models to the \textit{Lane-Boyes-Braem (LBB) II} model adapted by \citet{stungis} (Figure \ref{fig:tree}), where we color-coded the dendrogram such that similar handshapes share similar colors. \ky{TODO: add interesting observations? E.g., the stgcn vs. HD plot shows a nice separation between LBB2 clusters: STGCN can separate the closest clusters in LBB2 quite well, and handshape distance is small for compact pairs, medium for not broad pairs, high for broad pairs. }

\paragraph{Hierarchical clustering analysis.}
The dendrograms reveal distinct topological differences between models that corroborate the correlation results. The \textbf{LBB2 Model} organizes handshapes primarily by the `Compact' feature (whether the three middle fingers are closed). Within the `+Compact' group, it further distinguishes based on the number of bent fingers, while the `-Compact' group is organized by the number of extended fingers. Similarly, the \textbf{HD dendrogram} organizes handshapes by finger count, creating a primary split between handshapes with $\leq 1$ finger extended (with the exception of two handshapes that has the thumb and another finger extended) versus those with $>1$. It further sub-groups the first cluster by specific articulators (thumb vs. index extension) and the second by the count of extended fingers.

The \textbf{I3D-ASL dendrogram} diverges the most from the LBB2 Model. It primarily separates by handshapes that visually resemble a fist from those that do not. Within the `fist' cluster, it sorts by closure tightness. The non-fist cluster separates a group with extended thumbs but leaves the remaining handshapes in sub-clusters lacking clear patterns. This suggests I3D relies primarily on global visual texture (e.g., ``compact fist'' vs. ``open hand with protruding thumb'') rather than fine-grained articulatory features.

In contrast, the \textbf{STGCN-ASL dendrogram} closely recovers the structure of the LBB2 Model. It organizes handshapes into four primary clusters: (1) all three middle fingers are closed, (2) 4-5 extended fingers, (3) 2-3 extended fingers, and (4) just one extended finger (with the exception of the `K' handshape on the far right). STGCN also captures fine-grained sub-distinctions within these groups: in the first cluster, it distinguishes thumb/pinky extension; in the second, it separates curved fingers from straight extension. This structural alignment explains the STGCN's high correlation with human perception (LBB2) and the theoretical HD metric.

% This visualization highlights that human perception groups handshapes into intuitive clusters (e.g., compact fists vs. open palms), a structure that STGCN-ASL approximates more closely than I3D-ASL.

\section{Discussion \& Future Work}

Our investigation reveals that while phonological sensitivity emerges without supervision in SLR models, it is fractionated by architectural inductive biases: pose-based STGCN are more sensitive to fine-grained handshape contrasts, whereas pixel-based I3D are better at location distinctions. This functional split suggests that current training methods and architectures are insufficient for holistic sign language processing, and that a more complete model likely requires dual-stream integration of the skeletal abstraction captured by pose-based models and the holistic spatial processing captured by pixel-based models.

Furthermore, the strong correlation of STGCN and human perception serves as a benchmark for the structural integrity of the latent space in skeletal models. However, these representations remain brittle to domain shifts. The significant drop in phonological sensitivity on out-of-domain data such as Sem-Lex exposes a tendency to overfit production variables (e.g., recording conditions) rather than acquiring robust phonological abstractions. Thus, scaling current training paradigms on naturalistic data may not be sufficient to induce the robust phonological representations required for real-world applications.

While our framework for phonological sensitivity advances SLR evaluation, our study has three primary limitations. First, data sparsity restricted our analysis to handshape, location, and movement, omitting palm orientation and non-manual markers. Second, we used only supervised baselines due to the unavailability of open-source models; future work should investigate whether self-supervised objectives induce different representations. Finally, our focus on isolated signs leaves the effects of co-articulation and syntax in continuous SLR for future study.

Beyond these, our approach enables broader exploration of model robustness and generalizability. Future research can probe compositionality of model features, similarly to how word embeddings enable analogical reasoning, and explore multimodal architectures that bridge the trade-off between I3D and STGCN models. The minimal pairs could also be used to improve training itself: they supply hard negatives for contrastive fine-grained phonological discrimination, and could pre-train models to first acquire phonological distinctions that then transfer to lower-resource sign languages. Future work could also test for cross-linguistic transfer between sign languages to see whether phonological representations learned from higher-resource sign languages could generalize to lower-resources sign languages; testing transfer across structurally unrelated sign languages  could further isolate phonological constraints grounded in human body biomechanics from language-specific learning, which would inform data-efficient modeling.

\subsubsection*{Acknowledgements}
KY is supported by the Vitalik Buterin Ph.D. Fellowship in AI Existential Safety. AK is supported by NSF STEM-APWD 136073-5130403.

\bibliography{colm2026_conference}

@article{desai2024systemic,
  title={Systemic biases in sign language AI research: A deaf-led call to reevaluate research agendas},
  author={Desai, Aashaka and De Meulder, Maartje and Hochgesang, Julie A and Kocab, Annemarie and Lu, Alex X},
  journal={arXiv preprint arXiv:2403.02563},
  year={2024}
}

@inproceedings{gauthier2025emergent,
  title={Emergent morpho-phonological representations in self-supervised speech models},
  author={Gauthier, Jon and Breiss, Canaan and Leonard, Matthew K and Chang, Edward F},
  booktitle={Proceedings of the 2025 Conference on Empirical Methods in Natural Language Processing},
  pages={28055--28074},
  year={2025}
}

@inproceedings{yin-read-2020-better,
    title = "Better Sign Language Translation with {STMC}-Transformer",
    author = "Yin, Kayo  and
      Read, Jesse",
    editor = "Scott, Donia  and
      Bel, Nuria  and
      Zong, Chengqing",
    booktitle = "Proceedings of the 28th International Conference on Computational Linguistics",
    month = dec,
    year = "2020",
    address = "Barcelona, Spain (Online)",
    publisher = "International Committee on Computational Linguistics",
    url = "https://aclanthology.org/2020.coling-main.525/",
    doi = "10.18653/v1/2020.coling-main.525",
    pages = "5975--5989"
}

@article{desai2023asl,
  title={ASL Citizen: A Community-Sourced Dataset for Advancing Isolated Sign Language Recognition},
  author={Desai, Aashaka and Berger, Lauren and Minakov, Fyodor O and Milan, Vanessa and Singh, Chinmay and Pumphrey, Kriston and Ladner, Richard E and Daum{\'e} III, Hal and Lu, Alex X and Caselli, Naomi and Bragg, Danielle},
  journal={arXiv preprint arXiv:2304.05934},
  year={2023}
}

@article{kezar2023sem,
  title={The sem-lex benchmark: Modeling asl signs and their phonemes},
  author={Kezar, Lee and Thomason, Jesse and Caselli, Naomi and Sehyr, Zed and Pontecorvo, Elana},
  journal={Proceedings of the 25th International ACM SIGACCESS Conference on Computers and Accessibility},
  year={2023}
}

@article{liddell1989american,
  title={American sign language: The phonological base},
  author={Liddell, Scott K and Johnson, Robert E},
  journal={Sign language studies},
  volume={64},
  number={1},
  pages={195--277},
  year={1989},
  publisher={Gallaudet University Press}
}

@article{bilge2022towards,
  title={Towards zero-shot sign language recognition},
  author={Bilge, Yunus Can and Cinbis, Ramazan Gokberk and Ikizler-Cinbis, Nazli},
  journal={IEEE transactions on pattern analysis and machine intelligence},
  volume={45},
  number={1},
  pages={1217--1232},
  year={2022},
  publisher={IEEE}
}

@inproceedings{pasad2021layer,
  title={Layer-wise analysis of a self-supervised speech representation model},
  author={Pasad, Ankita and Chou, Ju-Chieh and Livescu, Karen},
  booktitle={2021 IEEE Automatic Speech Recognition and Understanding Workshop (ASRU)},
  pages={914--921},
  year={2021},
  organization={IEEE}
}

@article{belinkov2017analyzing,
  title={Analyzing hidden representations in end-to-end automatic speech recognition systems},
  author={Belinkov, Yonatan and Glass, James},
  journal={Advances in Neural Information Processing Systems},
  volume={30},
  year={2017}
}

@inproceedings{tavella2022phonology,
  title={Phonology recognition in american sign language},
  author={Tavella, Federico and Galata, Aphrodite and Cangelosi, Angelo},
  booktitle={ICASSP 2022-2022 IEEE International Conference on Acoustics, Speech and Signal Processing (ICASSP)},
  pages={8452--8456},
  year={2022},
  organization={IEEE}
}

@article{sandler2012phonological,
  title={The phonological organization of sign languages},
  author={Sandler, Wendy},
  journal={Language and linguistics compass},
  volume={6},
  number={3},
  pages={162--182},
  year={2012},
  publisher={Wiley Online Library}
}

@book{battison1978lexical,
    title={Lexical borrowing in American sign language.},
    author={Battison, Robbin},
    year={1978},
    publisher={ERIC}
}

@article{stokoe,
    author = {Stokoe, William C., Jr.},
    title = "{Sign Language Structure: An Outline of the Visual Communication Systems of the American Deaf}",
    journal = {The Journal of Deaf Studies and Deaf Education},
    volume = {10},
    number = {1},
    pages = {3-37},
    year = {1960},
    month = {01},
    issn = {1081-4159},
    doi = {10.1093/deafed/eni001},
    url = {https://doi.org/10.1093/deafed/eni001},
    eprint = {https://academic.oup.com/jdsde/article-pdf/10/1/3/1034248/eni001.pdf},
}

@inproceedings{carreira2017quo,
  title={Quo vadis, action recognition? a new model and the kinetics dataset},
  author={Carreira, Joao and Zisserman, Andrew},
  booktitle={proceedings of the IEEE Conference on Computer Vision and Pattern Recognition},
  pages={6299--6308},
  year={2017}
}

@inproceedings{yan2018spatial,
  title={Spatial temporal graph convolutional networks for skeleton-based action recognition},
  author={Yan, Sijie and Xiong, Yuanjun and Lin, Dahua},
  booktitle={Proceedings of the AAAI conference on artificial intelligence},
  volume={32},
  number={1},
  year={2018}
}

@article{guan2025mska,
  title={MSKA: Multi-stream keypoint attention network for sign language recognition and translation},
  author={Guan, Mo and Wang, Yan and Ma, Guangkun and Liu, Jiarui and Sun, Mingzu},
  journal={Pattern Recognition},
  volume={165},
  pages={111602},
  year={2025},
  publisher={Elsevier}
}

@inproceedings{camgoz2020sign,
  title={Sign language transformers: Joint end-to-end sign language recognition and translation},
  author={Camgoz, Necati Cihan and Koller, Oscar and Hadfield, Simon and Bowden, Richard},
  booktitle={Proceedings of the IEEE/CVF conference on computer vision and pattern recognition},
  pages={10023--10033},
  year={2020}
}

@article{asl-lex,
  title={The ASL-LEX 2.0 Project: A database of lexical and phonological properties for 2,723 signs in American Sign Language},
  author={Sehyr, Zed Sevcikova and Caselli, Naomi and Cohen-Goldberg, Ariel M and Emmorey, Karen},
  journal={The Journal of Deaf Studies and Deaf Education},
  volume={26},
  number={2},
  pages={263--277},
  year={2021},
  publisher={Oxford University Press}
}

@article{stungis,
  title={Identification and discrimination of handshape in American Sign Language},
  author={Stungis, James},
  journal={Perception \& Psychophysics},
  volume={29},
  number={3},
  pages={261--276},
  year={1981},
  publisher={Springer}
}

@unpublished{hcs,
  title  = {Title Redacted for Blind Review},
  author = {{Anonymous Authors}},
  year   = {2026},
  note   = {Manuscript in preparation},
}

@article{lane1976preliminaries,
  title={Preliminaries to a distinctive feature analysis of handshapes in American Sign Language},
  author={Lane, Harlan and Boyes-Braem, Penny and Bellugi, Ursula},
  journal={Cognitive Psychology},
  volume={8},
  number={2},
  pages={263--289},
  year={1976},
  publisher={Elsevier}
}

@inproceedings{yin24acl,
    title = {Pressures for Communicative Efficiency in American Sign Language},
    author = {Yin, Kayo and Regier, Terry and Klein, Dan},
    booktitle = {Annual Conference of the Association for Computational Linguistics (ACL)},
    month = {August},
    year = {2024}
}

@article{lieberman2020lexical,
  title={Lexical recognition in deaf children learning American Sign Language: Activation of semantic and phonological features of signs},
  author={Lieberman, Amy M and Borovsky, Arielle},
  journal={Language learning},
  volume={70},
  number={4},
  pages={935--973},
  year={2020},
  publisher={Wiley Online Library}
}

@article{williams2017operationalization,
  title={Operationalization of sign language phonological similarity and its effects on lexical access},
  author={Williams, Joshua T and Stone, Adam and Newman, Sharlene D},
  journal={The Journal of Deaf Studies and Deaf Education},
  volume={22},
  number={3},
  pages={303--315},
  year={2017},
  publisher={Oxford University Press}
}

@article{d1978u,
  title={U-statistic hierarchical clustering},
  author={D’Andrade, Roy G},
  journal={Psychometrika},
  volume={43},
  number={1},
  pages={59--67},
  year={1978},
  publisher={Springer-Verlag}
}

@article{meade2018phonological,
  title={Phonological and semantic priming in American Sign Language: N300 and N400 effects},
  author={Meade, Gabriela and Lee, Brittany and Midgley, Katherine J and Holcomb, Phillip J and Emmorey, Karen},
  journal={Language, Cognition and Neuroscience},
  volume={33},
  number={9},
  pages={1092--1106},
  year={2018},
  publisher={Taylor \& Francis}
}

@article{carreiras2008lexical,
  title={Lexical processing in Spanish sign language (LSE)},
  author={Carreiras, Manuel and Guti{\'e}rrez-Sigut, Eva and Baquero, Silvia and Corina, David},
  journal={Journal of Memory and Language},
  volume={58},
  number={1},
  pages={100--122},
  year={2008},
  publisher={Elsevier}
}
\bibliographystyle{colm2026_conference}

\appendix
\section{Appendix}

\subsection{SLR model input modalities}
\label{app:input}

\begin{figure}[h]
    \centering
    \includegraphics[width=0.7\linewidth]{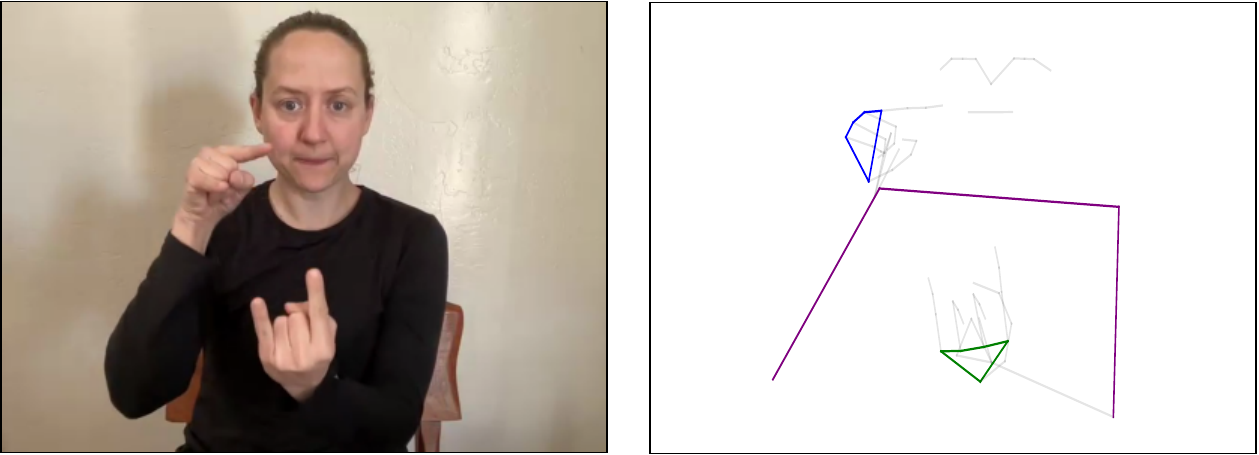}
\caption{\textbf{Comparison of model input modalities.} (Left) Raw RGB video frames processed by the I3D model. (Right) Skeletal pose graph estimation processed by the STGCN model. Both inputs depict a frame from the sign ``HIPPO.''}
\label{fig:mp}
\end{figure}

In Figure \ref{fig:mp}, we show examples of the video input modality to the I3D and STGCN models.

\subsection{Qualitative analysis of minimal pair sensitivity}
\label{app:quant}

In Table \ref{tab:qualitative}, we provide qualitative examples of \textit{t}-test results for I3D-ASL and STGCN-ASL models. The first three examples show cases where both models fail to represent the phonological change in minimal pairs of varying similarity as judged by humans. The first pair (\textit{HOSPITAL / PATIENT}) has a change in handshape, where the handshapes primarily differ by the position of the index finger. The second pair (\textit{5DOLLARS / FIFTH}) has a change in movement, where \textit{5DOLLARS} is signed by twisting the wrist once, while \textit{FIFTH} is signed by twisting the wrist inward-outward in small movements. The third pair (\textit{BEAVER / TABLE}) has a change in location, where the elbow of the dominant arm and wrist of the non-dominant arm stay in contact for \textit{BEAVER}, but neither elbows and wrists stay in contact during the movements for \textit{TABLE}. 

The final two examples show the divergence between architectures' phonological sensitivity. In example 4, only STGCN fails to distinguish \textit{FEEL / HAPPY}, where these signs mostly differ in whether the middle finger is bent or straight. In example \blue{5}, only I3D fails to distinguish \textit{SEVEN / NINE}, where the handshapes differ in the finger that is in contact with the thumb.

\begin{table}[h]
    \centering
    \caption{Qualitative examples of model sensitivity to minimal pairs in ASL Citizen.}
    \label{tab:qualitative}
    \begin{tabular}{C{1.5cm} C{3.5cm} C{3.5cm} C{1.5cm} C{1cm}  C{1.1cm}}
        \toprule
        \textbf{Minimal Pair} & \textbf{Sign 1} & \textbf{Sign 2} & \textbf{Human} & \textbf{I3D} & \textbf{STGCN} \\ 
        \midrule
        
        HOSPITAL / PATIENT & 
        \includegraphics[width=3.5cm]{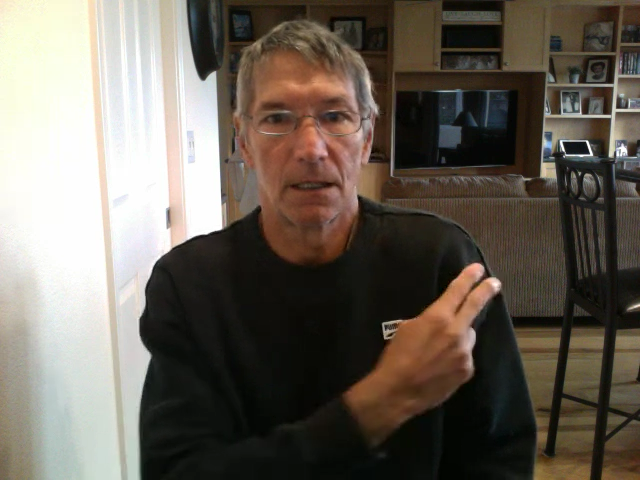} & 
        \includegraphics[width=3.5cm]{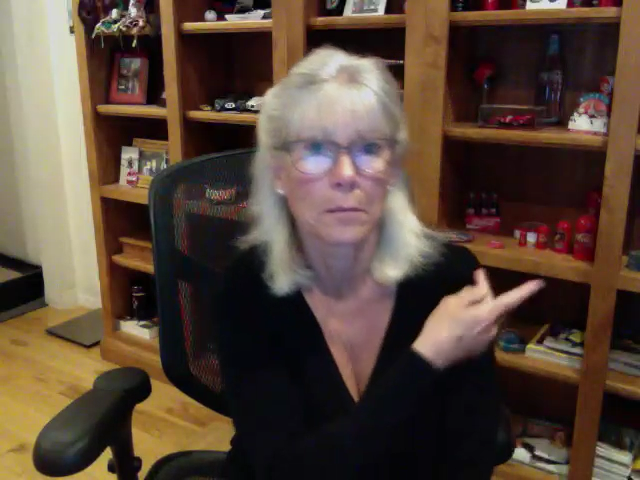} & 
        Very similar & 0.56 & -0.26 \\
        
        \addlinespace[10pt]
        
        5DOLLARS / FIFTH & 
        \includegraphics[width=3.5cm]{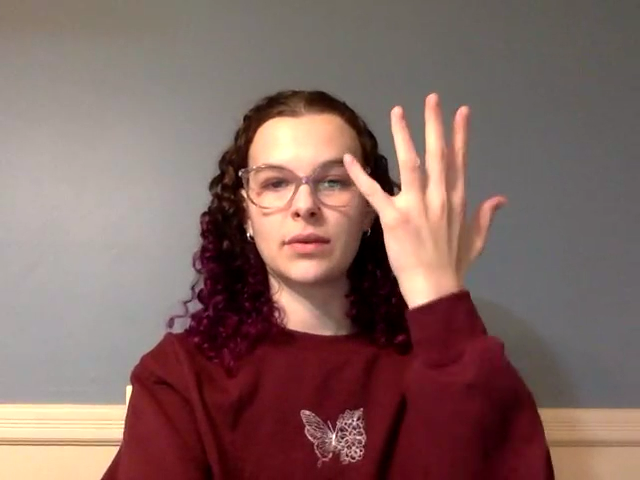} & 
       \includegraphics[width=3.5cm]{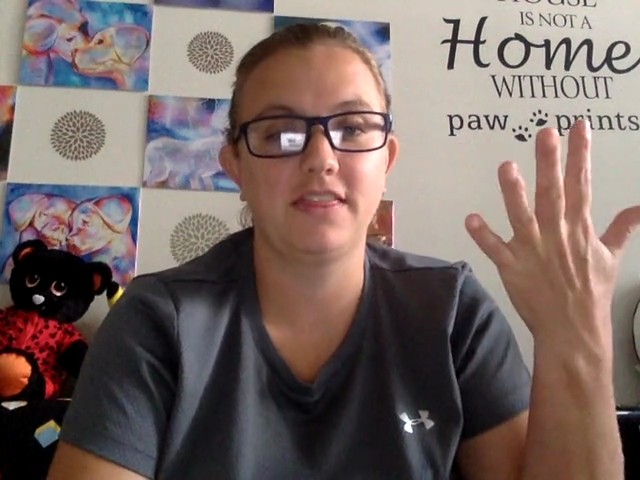} & 
        Somewhat similar & 0.10 & -1.01 \\

        \addlinespace[10pt]
        
        BEAVER / TABLE & 
        \includegraphics[width=3.5cm]{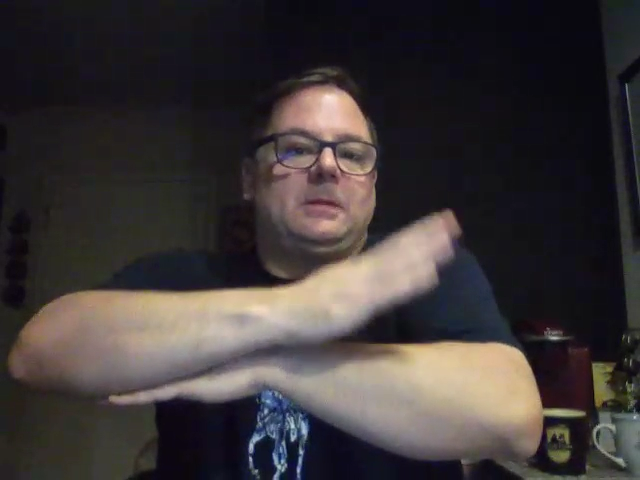} & 
       \includegraphics[width=3.5cm]{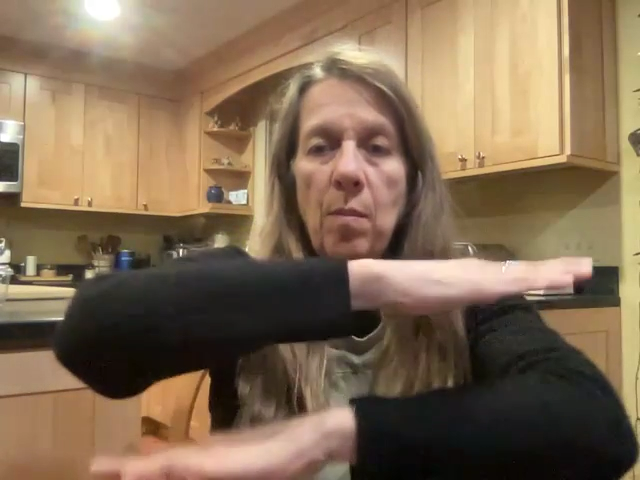} & 
        Not similar & 0.24 & -0.81 \\

        \addlinespace[10pt]
        
        FEEL / HAPPY & 
        \includegraphics[width=3.5cm]{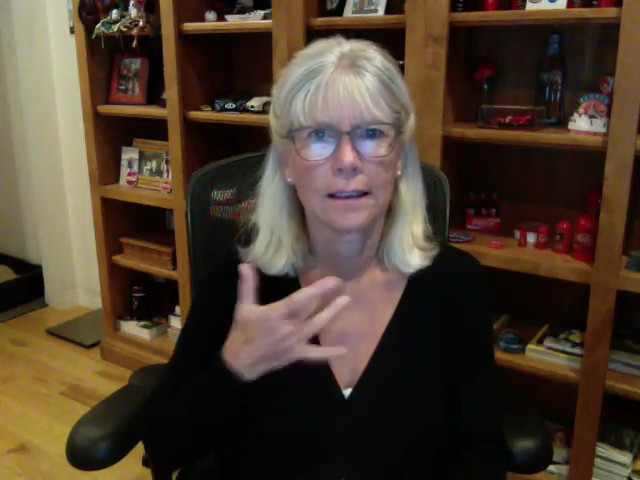} & 
       \includegraphics[width=3.5cm]{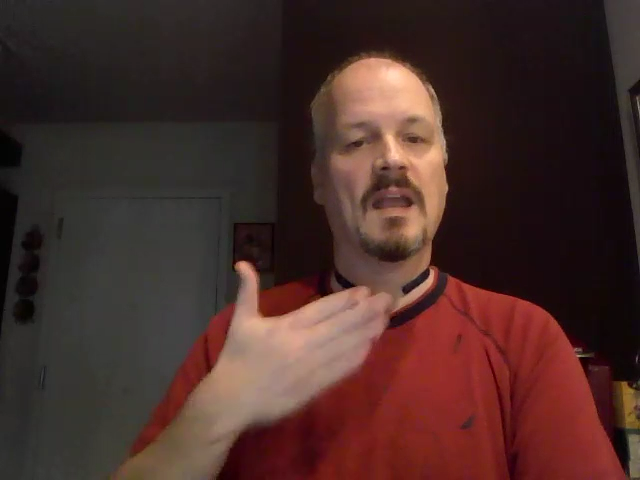} & 
        Somewhat similar & 4.89 & -1.26 \\

        \addlinespace[10pt]
        
        SEVEN / NINE & 
        \includegraphics[width=3.5cm]{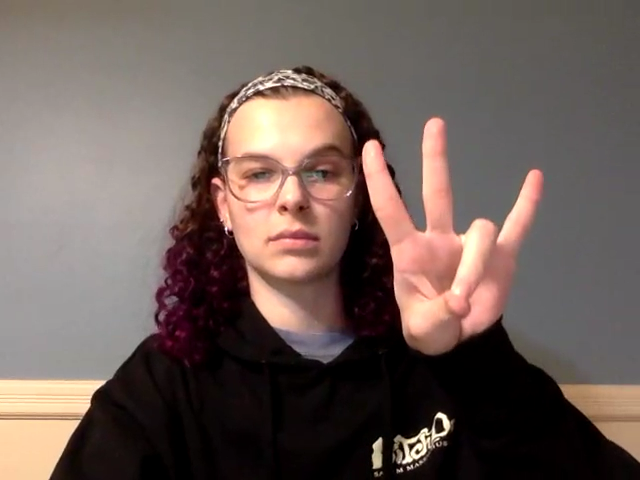} & 
       \includegraphics[width=3.5cm]{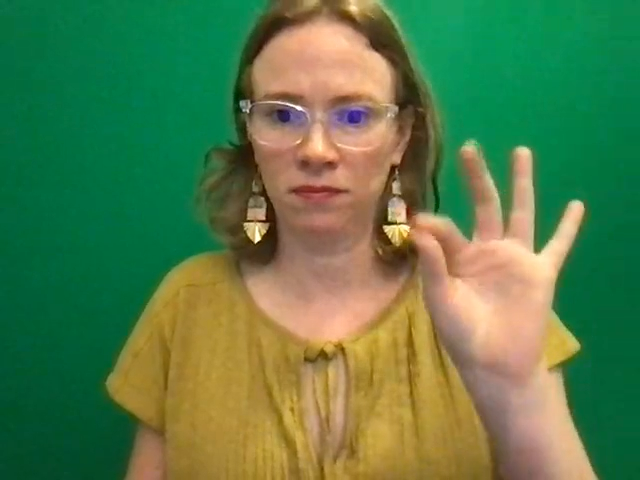} & 
        Not similar & -0.72 & 10.71 \\
        
        \bottomrule
    \end{tabular}
\end{table}

\end{document}